\documentclass[11pt]{article}

\PassOptionsToPackage{table}{xcolor}
\PassOptionsToPackage{hyperfootnotes=false}{hyperref}
\usepackage[preprint]{acl}

\usepackage{times}
\usepackage{latexsym}
\usepackage[T1]{fontenc}
\usepackage[utf8]{inputenc}
\usepackage{microtype}
\usepackage{inconsolata}
\usepackage{graphicx}
\usepackage{amsmath}
\usepackage{amssymb}
\usepackage{bm}
\usepackage{booktabs}
\usepackage{multirow}
\usepackage{array}
\usepackage{xcolor}
\usepackage{placeins}

\newcommand{\method}{\textsc{TVER}}
\newcommand{\ve}{\textsc{VE-25\%}}
\definecolor{tverrow}{HTML}{F1FAF5}
\definecolor{tverpurple}{HTML}{2E7D57}

\newcommand{\tvergain}[1]{\,{\scriptsize\textcolor{tverpurple}{(\(\uparrow\)#1)}}}
\newcommand{\tvernum}[2]{\textbf{\textcolor{tverpurple}{#1}}\tvergain{#2}}
\newcommand{\bestbase}[1]{\underline{#1}}

\newcommand{\apptablecaption}[2]{%
  \captionsetup{type=table,hypcap=false}%
  \caption{#2}%
  \label{#1}%
}

\title{The Image Still Has a Say:\\
Final-Answer Repair for Multimodal Chain-of-Thought}

\author{Yulan Pan\thanks{Corresponding author.} \quad Xiaoyu Xu \quad Zhihong Shen \\
Fuzhou University \\
\texttt{isyolaa@outlook.com}}

\hypersetup{pdftitle={The Image Still Has a Say: Final-Answer Repair for Multimodal Chain-of-Thought},pdfauthor={Yulan Pan, Xiaoyu Xu, Zhihong Shen}}

\begin{document}
\maketitle

\begin{abstract}
Although chain-of-thought prompting has improved multimodal reasoning, it is still difficult to produce correct final answers, as the reasoning can drift away from the image.
Besides, the trace is already fixed when the answer is generated, leaving no room for self-correction.
Considering the image is still available at the final-answer step, we think it is reasonable to fix the answer there without rewriting the reasoning.
As such, the problem becomes to correct the final answer using the image under fixed reasoning.
Unlike methods that improve trace generation through training, resampling, or decoding-time grounding, we repair the final-answer readout while keeping the reasoning trace fixed.
The motivation is based on our finding that a wrong option can become prematurely committed in the reasoning while the correct option still has stronger image support in middle-to-deep layers.
Thus, we propose Text-to-Visual Evidence Reallocation (TVER), a training-free method for final-answer repair.
Specifically, TVER selects answer-related text and question-matched image patches, and moves bounded attention from text to patches.
Furthermore, the patches are chosen in a middle-to-deep layer window based on question relevance.
Our method improves answers across benchmarks and backbones, with gains in both multiple-choice and open-answer evaluations.
\end{abstract}

\section{Introduction}

Chain-of-thought prompting asks models to write intermediate reasoning before answering \citep{wei2022cot}, and multimodal CoT extends this idea to visual reasoning \citep{lu2022scienceqa,zhang2024mmcot}.
Each generated step then conditions the next steps and the final answer.
When long multimodal reasoning drifts away from image-grounded content \citep{xu2025thinkingseeing}, the written text can become stronger than the remaining visual signal.
As a result, the final answer may follow misleading text in the trace, even though the image is still present and may still support another answer.

\begin{figure}[t]
\centering
\includegraphics[width=0.98\linewidth]{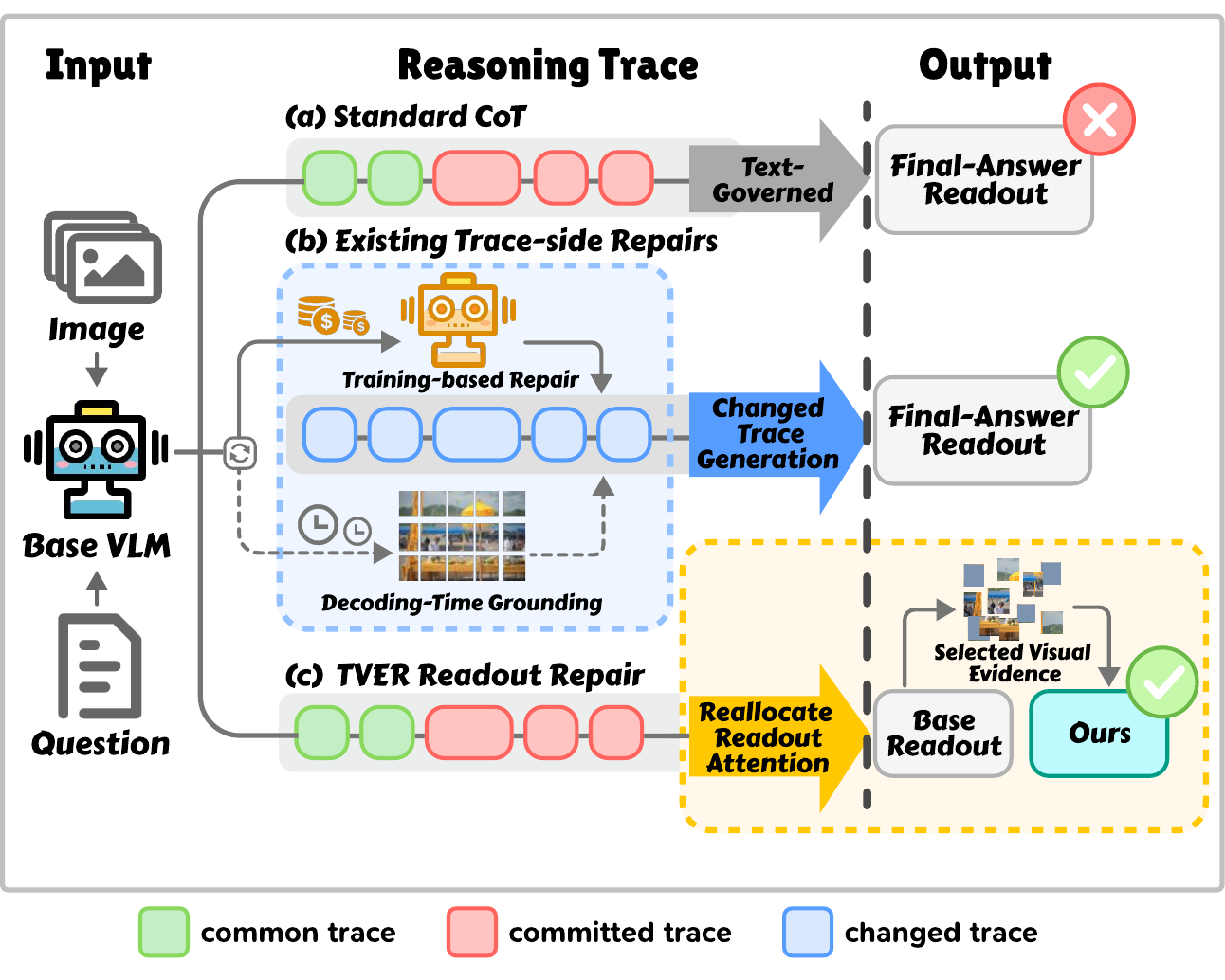}
\caption{The problem and our main idea. Prior training or decoding-time methods change the reasoning trace before the final answer is read. \method{} keeps the trace fixed and changes only the final-answer readout by reallocating attention from answer-related text in the fixed context to selected visual evidence.}
\label{fig:intervention-site}
\end{figure}

Intuitively, one may try to avoid this failure by training the model to write better grounded traces.
Post-training, grounded CoT supervision, and reinforcement-style optimization can improve multimodal reasoning behavior \citep{man2025argus,zhang2025improvevlmcot,liu2025visualrft,ni2025pointrft,huang2026visionr1}.
However, these methods require additional optimization and often task-specific supervision or rewards.
Besides, because they change the model or the trace-generation policy, they do not directly explain what should be done when a particular trace has already been written.

Consequently, work has also moved to inference time without weight updates.
One solution is to change generation itself, such as by resampling reasoning paths, adding visual search, or steering decoding with visual contrast and attention signals \citep{wang2023selfconsistency,wu2024vstar,leng2024vcd,chen2024halc,liu2024pai,kang2025var,li2025vista,tu2026attnreal}.
Another solution is to revise the completed response after decoding \citep{zhou2024lure,yin2024woodpecker}.
Nevertheless, these methods still act during generation or after the response is written.
They do not isolate the final-answer readout itself, where the written trace may already favor one answer while the image still supports another.

Thinking beyond trace generation, we propose to explicitly investigate the final-answer readout under a fixed reasoning trace 1) to avoid treating the whole response as a generation problem; 2) without updating model weights, regenerating the trace, or editing the trace.
We assume that an answer-line error can involve a text-side commitment from the written trace and remaining visual evidence from the image.
That is, the trace may already favor one answer, while the image may still support another at readout.
If these two factors can be separated at the answer line, then final-answer repair can be studied without rewriting the reasoning.
To test this possibility, we examine the final-answer readout of open-source LVLMs, focusing on hidden states and attention patterns.
We begin with the following diagnostics.

\section{Analysis and Motivation}
\label{sec:motivation}

\begin{figure}[t]
\centering
\includegraphics[width=0.98\linewidth]{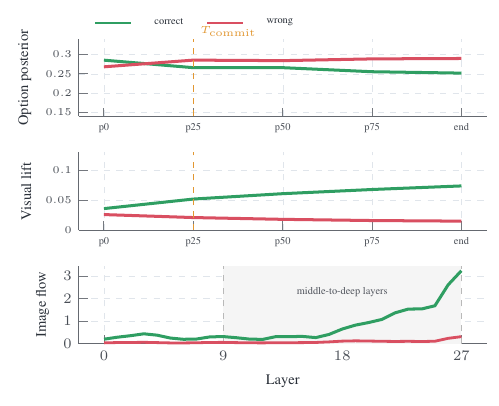}
\caption{Motivating diagnostic for final-answer repair. The wrong option becomes stable before the answer line, while the correct option keeps larger visual lift and stronger image-to-answer flow in middle-to-deep layers.}
\label{fig:diagnostic-trace}
\end{figure}

\paragraph{A fixed trace may settle on a wrong answer too early.}
We first investigate when the model commits to a wrong option during reasoning, i.e., we want to know how early the wrong option selected in the final answer becomes stable.
To this end, we first let the model generate the complete CoT and final answer and retain the multiple-choice examples answered incorrectly.
Specifically, keeping the image and question fixed, we score each candidate option at p0, p25, p50, p75, and p100.
Here, pX means the first X\% of the CoT; at each pX, we normalize the raw scores \(s(o)\) across candidate set \(\mathcal{O}\) to obtain option scores \(P(o)\):
\begin{equation}
P(o)=
\frac{\exp s(o)}
{\sum_{o'\in\mathcal{O}}\exp s(o')}.
\end{equation}
For each error, we define \(T_{\mathrm{commit}}\) as the first checkpoint at which the wrong option selected in the final answer becomes top-1 and remains top-1 at every subsequent checkpoint.
When \(T_{\mathrm{commit}}\) is at or before p50, the model has settled on the wrong option by the midpoint of its reasoning, which we regard as premature commitment.
Then, we find that premature commitment occurs in 33.8\% of CoT errors on average across the five multiple-choice pools for Qwen2.5-VL-7B.
At p0 in Figure~\ref{fig:diagnostic-trace}, the correct option is slightly ahead of the wrong option (0.285 vs. 0.268); by p25, the ranking reverses, and the wrong option becomes top-1 (0.285 vs. 0.266).
However, the remaining reasoning does not change this ranking: the wrong option stays top-1 until the end of the CoT.
That is, the model settles on the wrong option after only the first quarter of its reasoning, and the subsequent reasoning does not correct this choice.

\paragraph{The image may still favor the correct option.}
Then, we investigate whether the image still supports the correct option when the CoT ranks the wrong option first.
To this end, following visual-contrast work \citep{leng2024vcd}, we replace the original image with a blank image, a blurred image, or an unrelated image, while keeping the question, options, and the CoT at each pX unchanged.
Specifically, at each pX, we calculate the score \(P(o; I')\) of every option \(o\) under each image \(I'\) and define visual lift \(V(o)\), with \(\mathcal{C}\) denoting the three replacement images:
\begin{equation}
V(o)=P(o; I)
-\frac{1}{|\mathcal{C}|}\sum_{\tilde I\in\mathcal{C}}P(o; \tilde I).
\end{equation}
We use a positive visual lift to indicate that the original image supports an option more strongly than the replacement images.
Compared with p0, the visual lift of the correct option increases from 0.036 to 0.074 at p100, while that of the wrong option decreases from 0.026 to 0.015.
However, the wrong option remains top-1 at p100, although the original image provides stronger support for the correct option.
That is, the CoT favors the wrong option, while the image still supports the correct option more strongly.

\paragraph{Visual evidence can still reach the readout.}
Then, we ask whether this image-side signal can reach the final-answer tokens.
Visual lift alone would not help if the answer computation no longer used image tokens.
We estimate the path for each option \(o\) by scoring the option line and aggregating an attention-gradient attribution score from final-answer tokens to image tokens \citep{chefer2021generic}:
\begin{equation}
F_t^\ell(o)=
\sum_{q\in \mathcal{A}_o}\sum_{k\in \mathcal{K}_I}
\left|a^\ell_{qk}\frac{\partial s(o)}{\partial a^\ell_{qk}}\right|.
\end{equation}
Here \(\mathcal{A}_o\) is the answer-token set for option \(o\), and \(\mathcal{K}_I\) is the image-token set.
The lift asks whether changing the image changes the score; the flow asks whether final-answer scoring uses image tokens.
In Figure~\ref{fig:diagnostic-trace}, the flow is also larger for the correct option in the middle-to-deep layers.
Averaged over this window, the image-to-answer flow is 0.954 for the correct option and 0.099 for the wrong option.
The three measurements agree: the prefix posterior can favor the wrong answer, while visual support and image-to-answer flow remain stronger for the correct one.
Thus, the text-side commitment and image-side evidence can be divided at readout: one can dominate the posterior, while the other still reaches the answer computation.
This is the situation a readout repair should target.

Based on this finding, we propose Text-to-Visual Evidence Reallocation (\method{}), a training-free method for final-answer repair under a fixed multimodal CoT.
As shown in Figure~\ref{fig:intervention-site}, \method{} keeps the written trace unchanged and treats the final answer as a readout problem rather than a trace-generation problem.
It identifies the text-side context that pulls the answer toward the current readout and the visual evidence that is still relevant to the question, then gives this evidence more influence at the answer line.
In this way, the method targets the conflict exposed by the diagnostics without producing a new reasoning path.

We summarize our contributions as follows.
\textbf{1)} We define the fixed-trace final-answer repair setting for multimodal CoT, where the reasoning has already been written and only the final answer is repaired.
\textbf{2)} We provide diagnostics showing that a wrong text-side commitment can coexist with stronger visual support and image-to-answer flow for the correct option.
\textbf{3)} We introduce \method{}, a plug-and-play readout intervention that improves answers across benchmarks and open-source backbones, including both multiple-choice and open-answer evaluation.

\section{Method}
\label{sec:method}

\begin{figure*}[t]
\centering
\includegraphics[width=0.98\textwidth]{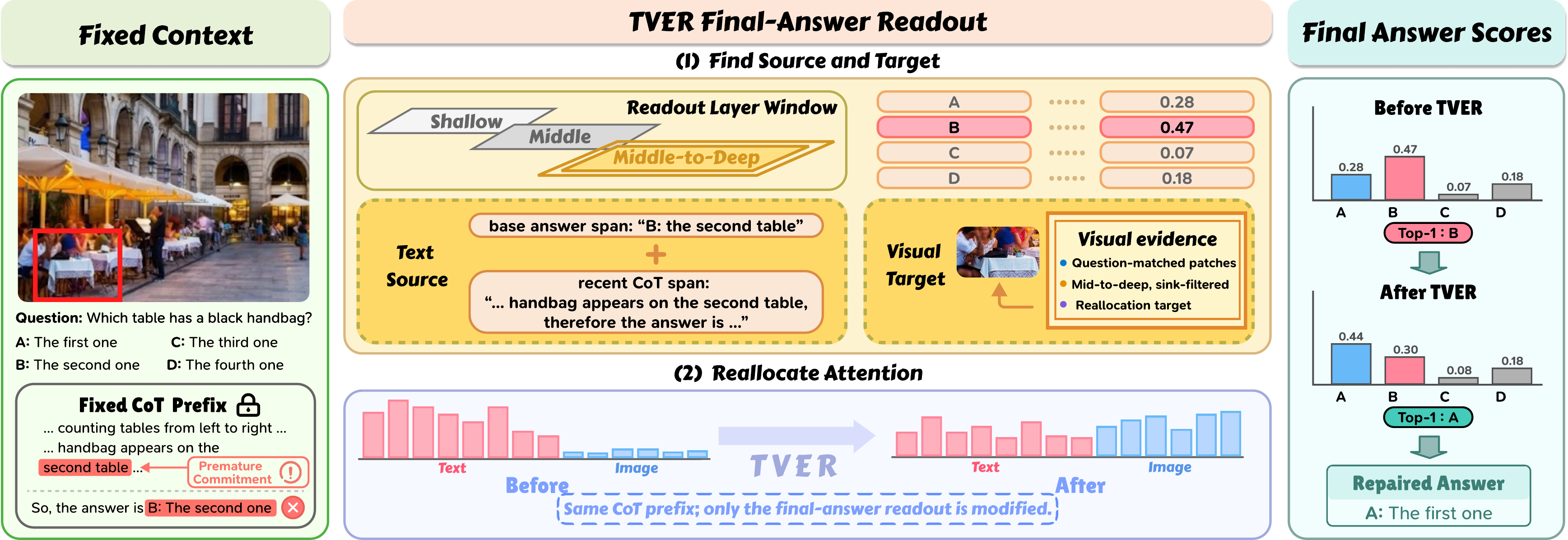}
\caption{Overview of \method{}. The image, question, options, and CoT prefix are fixed. \method{} selects a text-side source and a visual evidence target from the unmodified readout, then reruns only the final-answer line with capped attention reallocation.}
\label{fig:method-operator}
\end{figure*}

Suppose that a multimodal CoT has already been written for a given image and question.
Our goal is to repair the final answer without rewriting the CoT.
Motivated by the diagnostics in Section~\ref{sec:motivation}, \method{} uses the unmodified readout to identify the answer-related text and question-specific visual evidence.
Figure~\ref{fig:method-operator} depicts the framework of \method{}.
It includes three stages.
We first record the unmodified fixed-trace readout, which provides the base answer, hidden states, and attention maps.
Then, we select the answer-related text source and question-specific visual target.
After selecting the source and target, we rerun only the final-answer line with capped evidence reallocation.

\subsection{Fixed-Trace Readout}
\label{sec:method-readout}

Given the fixed CoT prefix, \method{} first runs the ordinary final-answer readout and obtains the base answer, denoted as \(a_{\mathrm{base}}\).
This pass is unmodified; it records the model's current answer and provides the hidden states and attention maps used for source and target selection.
For multiple-choice examples, we extract the option from \(a_{\mathrm{base}}\) using the same answer extractor as evaluation.
For open-answer examples, we keep \(a_{\mathrm{base}}\) as the generated answer string and do not extract an option span.

\subsection{Source and Target Selection}
\label{sec:method-selection}

The base answer \(a_{\mathrm{base}}\) records the model's current readout.
We then select a text-side source \(\mathcal{S}\) and a visual target \(\mathcal{V}\) before the reallocation step.

\paragraph{Text-side source.}
As shown in Section~\ref{sec:motivation}, the fixed prefix can already favor a wrong option, and the final readout tends to follow answer-related text in the fixed context.
We therefore take the option selected by the unmodified readout and the recent reasoning that leads into the answer line as the text-side source.
For multiple-choice examples, let \(o_{\mathrm{base}}\) be the option extracted from \(a_{\mathrm{base}}\).
We define \(\mathcal{T}_{\mathrm{ans}}\) as the spans tied to \(o_{\mathrm{base}}\): its label and option text in the prompt, plus explicit mentions in the generated reasoning.
\begin{equation}
\mathcal{S}=\mathcal{T}_{\mathrm{ans}}\cup \mathcal{R}_{\mathrm{recent}},
\end{equation}
where \(\mathcal{R}_{\mathrm{recent}}\) is the last non-empty segment before the answer marker after splitting the generated reasoning at newlines and sentence-final punctuation.
For open-answer examples, there is no option span, so we use \(\mathcal{S}=\mathcal{R}_{\mathrm{recent}}\).

\paragraph{Visual target.}
We next choose the image tokens that receive the reallocated attention.
As shown in Figure~\ref{fig:diagnostic-trace}, the image-to-answer flow separates the correct and wrong options mainly in the middle-to-deep layers.
We therefore use a middle-to-deep layer window, denoted by \(\mathcal{L}\), for both target selection and the reallocation step below.
For each backbone, \(\mathcal{L}\) contains the final two thirds of its transformer layers.
The target should also be specific to the current question.
If the moved attention is spread over all image patches, much of it can fall on background or regions unrelated to the question.
For this reason, we score each patch by a layer-averaged visual-evidence score.

In the unmodified forward pass, let \(\mathcal{K}_I\) be the image patch tokens and \(z_i^\ell\) be the hidden state of token \(i\) at layer \(\ell\).
Let \(\mathcal{Q}_{\mathrm{qo}}\) contain the question tokens and, when options are available, the option tokens.
Let \(\mathcal{Q}_{\mathrm{text}}\) collect the text side of the fixed prefix, including the question, the options when present, and the generated reasoning tokens.
For each layer \(\ell\in\mathcal{L}\), we compute two patch scores: question--option relevance and broad text attention:
\begin{equation}
\begin{aligned}
u^\ell(k) &= \cos\!\left(z_k^\ell,\,
\operatorname*{avg}_{q\in\mathcal{Q}_{\mathrm{qo}}}z_q^\ell\right),\\
b^\ell(k) &=
\operatorname*{avg}_{\substack{h\in\mathcal{H}\\q\in\mathcal{Q}_{\mathrm{text}}}}
A^{\ell,h}_{q,k},
\end{aligned}
\end{equation}
where \(A^{\ell,h}_{q,k}\) is the attention weight from text token \(q\) to image token \(k\), and \(\mathcal{H}\) is the set of attention heads.
We subtract the broad-attention score because patches watched by much of the text are often common visual anchors, not the specific evidence needed here.
We rank-normalize the two terms within each layer and then average across layers:
\begin{equation}
\begin{aligned}
r^\ell(k) &=
\operatorname{rn}_{\mathcal{K}_I}(u^\ell(k))
-
\operatorname{rn}_{\mathcal{K}_I}(b^\ell(k)),\\
r(k) &= \operatorname*{avg}_{\ell\in\mathcal{L}} r^\ell(k).
\end{aligned}
\end{equation}
Here \(\operatorname{rn}_{\mathcal{K}_I}\) maps the lowest raw score in a layer to 0 and the highest to 1, using average ranks for ties.
We take the top 25\% image tokens under \(r(k)\) as \(\mathcal{V}\).
This selected \(\ve{}\) set is the visual target used by the reallocation step.

\subsection{Evidence Reallocation}
\label{sec:method-reallocation}

\method{} changes the final-answer readout with a source-to-target operation.
The written trace stays fixed and remains available as context; only the attention rows for new answer tokens are modified.
The operation moves part of the selected source attention to the selected visual target.
We apply the operation in the same middle-to-deep layer window \(\mathcal{L}\).

For each layer \(\ell\in\mathcal{L}\), head \(h\), and new answer token, let \(a\) be its post-softmax attention distribution.
The attention mass on the selected text-side source is
\begin{equation}
M_{\mathcal{S}}=\sum_{j\in\mathcal{S}}a_j.
\end{equation}
Rather than removing all source attention, \method{} moves only a capped amount:
\begin{equation}
m=\min(\beta M_{\mathcal{S}},\tau_{\max},M_{\mathcal{S}}).
\end{equation}
Here \(\beta\) controls the released fraction of source mass, and \(\tau_{\max}\) caps the absolute transferred mass.
When \(M_{\mathcal{S}}>0\), we set \(\gamma=m/M_{\mathcal{S}}\) and scale the source-token weights as \(a'_j=(1-\gamma)a_j\) for \(j\in\mathcal{S}\).
The moved mass is added to visual target tokens according to their visual-evidence scores:
\begin{equation}
\begin{aligned}
w_k &= \frac{\exp r(k)}{\sum_{u\in\mathcal{V}}\exp r(u)},\\
a'_k &= a_k+mw_k,\quad k\in\mathcal{V}.
\end{aligned}
\end{equation}
Tokens outside \(\mathcal{S}\cup\mathcal{V}\) keep their original attention weights, and each attention row remains normalized.
The modified row is then passed through the model's standard value aggregation, and decoding continues only for the final answer line.

\section{Experiments}

\subsection{Setups}

We evaluate whether the final answer can be repaired after the CoT trace has already been written.
For each example and backbone, the model first writes a CoT, and we keep the prefix before the answer marker fixed.
The CoT baseline keeps the original final answer, the re-answer control decodes another answer line from the same prefix, and \method{} decodes that line with reallocation applied.
All rows use the same answer parser or judge.
We tune \method{} only on TreeBench tuning rows with Qwen2.5-VL-7B, then freeze the setting for all reported results.
\method{} requires no weight updates.

\subsection{Benchmarks and Base Models}

\paragraph{Benchmarks.}
We use three groups of benchmarks.
TreeBench probes ``thinking with images'' with traceable visual evidence and relational reasoning \citep{wang2026treebench}.
V*Bench stresses visual search in high-resolution, crowded scenes \citep{wu2024vstar}.
These two benchmarks are closest to our setting, since many errors depend on concrete visual evidence.
We also test broader multimodal reasoning with MathVista \citep{lu2024mathvista} and MMStar \citep{chen2024mmstar}, following the move from perception-heavy LVLM evaluation \citep{fu2025mme} to expert multimodal reasoning \citep{yue2024mmmu}.
All four benchmarks are evaluated by answer accuracy.
We also use RH-Reason in the open-answer RH-Bench setting, so the evaluation is not limited to multiple-choice answers \citep{xu2025thinkingseeing}.

\paragraph{Base models.}
We test whether the same frozen setting transfers across model families and sizes.
The evaluation covers six backbones from three model families, spanning 7B to 72B parameters: Qwen2.5-VL (7B and 32B) \citep{bai2025qwen25vl}, InternVL3 (8B and 38B) \citep{zhu2025internvl3}, and LLaVA-OneVision (7B and 72B) \citep{li2024llavaonevision}, without finetuning any backbone.

\subsection{Main Results and Analysis}

\begin{table*}[!t]
\centering
\small
\setlength{\tabcolsep}{3.0pt}
\renewcommand{\arraystretch}{1.0}
\begin{tabular}{@{}p{0.225\textwidth}|>{\centering\arraybackslash}p{0.135\textwidth}>{\centering\arraybackslash}p{0.135\textwidth}|>{\centering\arraybackslash}p{0.135\textwidth}>{\centering\arraybackslash}p{0.135\textwidth}|>{\centering\arraybackslash}p{0.16\textwidth}@{}}
\toprule
Method & \multicolumn{2}{c|}{Visual Grounding} & \multicolumn{2}{c|}{General VQA} & Reasoning \\
\cmidrule(lr){2-3}\cmidrule(lr){4-5}\cmidrule(lr){6-6}
 & TreeBench$\uparrow$ & V*Bench$\uparrow$ & MathVista$\uparrow$ & MMStar$\uparrow$ & RH-Reason$\uparrow$ \\
\midrule
\rowcolor{black!4}
Qwen2.5-VL-7B & 37.0 & 76.4 & 68.2 & 63.9 & 39.7 \\
\quad + VCD & 40.5 & 78.0 & 69.0 & 65.0 & 41.3 \\
\quad + PAI & 38.3 & 77.0 & 68.5 & 64.4 & \bestbase{42.3} \\
\quad + VAR & 39.5 & 78.5 & 68.9 & 64.8 & 41.3 \\
\quad + VISTA & 41.0 & \bestbase{79.1} & 69.4 & 65.4 & 42.0 \\
\quad + ViGaL & \bestbase{41.2} & 78.5 & \bestbase{71.9} & 65.4 & 42.0 \\
\quad + Woodpecker & 39.5 & 78.5 & 69.4 & \bestbase{65.5} & 41.7 \\
\rowcolor{tverrow}
\quad + \textbf{\textcolor{tverpurple}{\method{}}} & \tvernum{41.7}{4.7} & \tvernum{79.6}{3.2} & \tvernum{70.0}{1.8} & \tvernum{66.5}{2.6} & \tvernum{42.7}{3.0} \\
\midrule
\rowcolor{black!4}
Qwen2.5-VL-32B & 42.5 & 81.2 & 74.7 & 69.5 & 47.0 \\
\quad + VCD & 43.7 & 81.7 & 75.0 & 69.9 & 47.7 \\
\quad + PAI & 43.2 & 81.2 & 74.9 & 69.8 & 48.0 \\
\quad + VAR & 44.0 & \bestbase{82.2} & 75.2 & \bestbase{70.5} & 48.3 \\
\quad + VISTA & \bestbase{44.2} & \bestbase{82.2} & \bestbase{75.5} & 70.4 & \bestbase{48.7} \\
\rowcolor{tverrow}
\quad + \textbf{\textcolor{tverpurple}{\method{}}} & \tvernum{44.9}{2.4} & \tvernum{82.2}{1.0} & \tvernum{76.1}{1.4} & \tvernum{70.9}{1.4} & \tvernum{49.3}{2.3} \\
\midrule
\rowcolor{black!4}
InternVL3-8B & 38.8 & 72.3 & 71.6 & 68.2 & 42.7 \\
\quad + VCD & 40.5 & 73.8 & 71.9 & 68.8 & 43.7 \\
\quad + PAI & 40.0 & 73.3 & 71.8 & 68.6 & 44.0 \\
\quad + VAR & 41.0 & 74.9 & 72.2 & 69.0 & 44.3 \\
\quad + VISTA & \bestbase{41.7} & \bestbase{75.4} & 72.8 & \bestbase{69.7} & \bestbase{44.7} \\
\quad + Woodpecker & 41.0 & 74.3 & \bestbase{72.9} & 69.6 & 44.3 \\
\rowcolor{tverrow}
\quad + \textbf{\textcolor{tverpurple}{\method{}}} & \tvernum{42.7}{3.9} & \tvernum{75.9}{3.7} & \tvernum{73.6}{2.0} & \tvernum{70.4}{2.2} & \tvernum{45.3}{2.6} \\
\midrule
\rowcolor{black!4}
InternVL3-38B & 42.0 & 79.6 & 75.1 & 71.5 & 47.3 \\
\quad + VCD & 43.0 & 80.1 & 75.3 & 71.9 & 48.0 \\
\quad + PAI & 42.7 & 79.1 & 75.2 & 71.8 & 48.3 \\
\quad + VAR & 43.2 & 80.6 & 75.4 & \bestbase{72.3} & 48.3 \\
\quad + VISTA & \bestbase{43.7} & \bestbase{81.2} & \bestbase{75.7} & 72.2 & \bestbase{48.7} \\
\rowcolor{tverrow}
\quad + \textbf{\textcolor{tverpurple}{\method{}}} & \tvernum{44.9}{2.9} & \tvernum{81.7}{2.1} & \tvernum{76.2}{1.1} & \tvernum{72.8}{1.3} & \tvernum{49.7}{2.4} \\
\midrule
\rowcolor{black!4}
LLaVA-OneVision-7B & 37.3 & 70.7 & 63.2 & 61.7 & 39.7 \\
\quad + VCD & 39.0 & 72.3 & 63.6 & 62.4 & 41.0 \\
\quad + PAI & 39.8 & 72.8 & 64.0 & \bestbase{63.5} & 41.3 \\
\quad + VAR & 39.5 & 73.3 & 63.9 & 62.6 & 41.3 \\
\quad + VISTA & \bestbase{40.7} & \bestbase{73.8} & \bestbase{64.6} & 63.4 & \bestbase{42.0} \\
\rowcolor{tverrow}
\quad + \textbf{\textcolor{tverpurple}{\method{}}} & \tvernum{41.7}{4.4} & \tvernum{74.3}{3.6} & \tvernum{65.1}{1.9} & \tvernum{64.2}{2.5} & \tvernum{42.7}{3.0} \\
\midrule
\rowcolor{black!4}
LLaVA-OneVision-72B & 40.5 & 73.8 & 66.5 & 65.2 & 43.7 \\
\quad + VCD & 41.7 & 74.9 & 66.9 & 65.9 & 45.0 \\
\quad + PAI & 41.5 & 75.4 & \bestbase{67.7} & 66.2 & \bestbase{46.3} \\
\quad + VAR & 42.2 & 75.9 & 67.2 & 66.4 & 45.3 \\
\quad + VISTA & \bestbase{43.0} & \bestbase{76.4} & 67.6 & \bestbase{66.8} & \bestbase{46.3} \\
\rowcolor{tverrow}
\quad + \textbf{\textcolor{tverpurple}{\method{}}} & \tvernum{44.0}{3.5} & \tvernum{77.0}{3.2} & \tvernum{68.1}{1.6} & \tvernum{67.3}{2.1} & \tvernum{47.0}{3.3} \\
\bottomrule
\end{tabular}
\caption{Performance comparison under the fixed-prefix final-answer protocol. All rows use the same answer extraction or judge, and \method{} uses one fixed setting across all model-task cells. Highlighted rows are \method{}; underlines mark the strongest non-\method{} result in each model-task group; arrows show gains over CoT, computed from unrounded scores.}
\label{tab:main-results}
\end{table*}

\paragraph{Main results.}
\method{} improves over CoT in all 30 model--task cells, with an average gain of +2.6 points (95\% paired-bootstrap CI: [+2.17, +2.98]).
On Qwen2.5-VL-7B, wrong-to-correct changes outnumber correct-to-wrong changes on all five benchmarks, with repair rates of 7.2--17.8\% and damage rates of 1.4--3.4\%.
Appendix Table~\ref{tab:repair-damage} reports the full results.

\method{} yields consistent gains on TreeBench and V*Bench, where many answers rely on localized visual evidence.
MathVista and MMStar also improve, showing that the effect extends beyond visual-grounding benchmarks.
On RH-Reason, the average gain across backbones is +2.8 points.
Since RH-Reason is judged by answer equivalence rather than option matching, this result shows that \method{} also improves open-answer responses.

Table~\ref{tab:main-results} also compares \method{} with four training-free inference baselines: VCD \citep{leng2024vcd}, PAI \citep{liu2024pai}, VAR \citep{kang2025var}, and VISTA \citep{li2025vista}, each run with the authors' recommended settings.
For each model and benchmark, we compare against the strongest of these four controls.
\method{} matches or exceeds this best training-free control in every case and is +0.7 points higher on average.
This suggests that source-to-visual readout reallocation adds value beyond generic visual contrast, image-attention reweighting, or generation-time steering.
Beyond these generation-time controls, we also compare with ViGaL \citep{xie2026vigal}, a post-trained model, and Woodpecker \citep{yin2024woodpecker}, a post-decoding revision method.
\method{} achieves comparable results across both comparisons without additional training.

\begin{figure}[t]
\centering
\includegraphics[width=\linewidth]{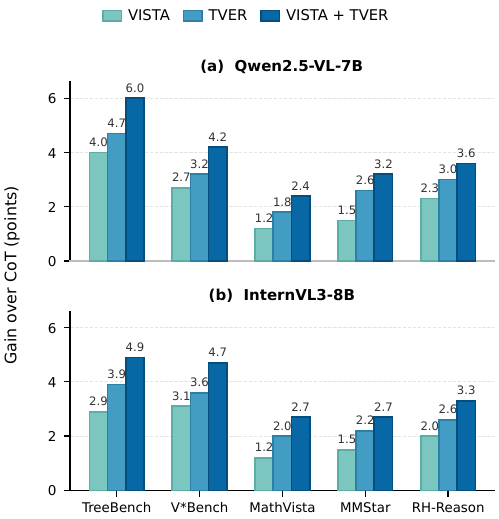}
\caption{Composability of VISTA and \method{} on Qwen2.5-VL-7B and InternVL3-8B. The x-axis lists the five benchmarks, and the y-axis shows accuracy gains over CoT in percentage points. VISTA and VISTA+\method{} share the same VISTA-generated trace and differ only at the final-answer readout; the combination gives higher gains than either method alone in all ten settings.}
\label{fig:vista-tver}
\end{figure}

\paragraph{Mechanism analysis.}
On CoT errors, the correct option has stronger hidden-state visual lift than the selected wrong option in every pool (+.057 on average; Appendix Table~\ref{tab:hidden-lift-appendix}).
After \method{}, the correct option's margin over the strongest wrong option increases in all five pools under the same fixed CoT (+.066 on average; Appendix Table~\ref{tab:closedset-appendix}), showing that the remaining visual support has greater influence on the final readout.

Figure~\ref{fig:vista-tver} further shows that combining \method{} with VISTA gives higher gains than either method alone on all five benchmarks for both backbones, indicating that final-answer repair complements generation-time steering.

\subsection{Ablations, Efficiency, and Robustness}

\paragraph{Module ablations.}
Table~\ref{tab:module-ablation} studies the source and target choices in \method{}.
The first rows compare simple controls.
Re-answering the same trace gives only +0.4 points.
Downweighting answer text or releasing attention without a visual target also gives smaller gains than full reallocation.
This shows that \method{} does not work by merely asking again or weakening text cues; the released attention needs a selected visual target.

\begin{table}[!t]
\centering
\small
\setlength{\tabcolsep}{2.7pt}
\begin{tabular}{llcc}
\toprule
Part & Setting & Acc. & \(\Delta\) \\
\midrule
\rowcolor{black!4}
Base & CoT & 50.6 & -- \\
\midrule
Control & re-answer & 51.0 & +0.4 \\
Control & answer-text downweight & 51.6 & +1.0 \\
Control & release-only & 52.0 & +1.4 \\
\midrule
Source & recent-only & 51.8 & +1.2 \\
Source & answer text & 52.6 & +2.0 \\
\rowcolor{tverrow}
Source & \textbf{\textcolor{tverpurple}{answer text+recent}} & \textbf{\textcolor{tverpurple}{53.8}} & \textbf{\textcolor{tverpurple}{+3.2}} \\
\midrule
Target & random 25\% & 51.0 & +0.4 \\
Target & all-image & 52.2 & +1.6 \\
Target & task-only & 52.9 & +2.3 \\
\rowcolor{tverrow}
Target & \textbf{\textcolor{tverpurple}{\ve{}}} & \textbf{\textcolor{tverpurple}{53.8}} & \textbf{\textcolor{tverpurple}{+3.2}} \\
\bottomrule
\end{tabular}
\caption{Module ablations on matched MCQ rows. Source rows fix the target as \ve{}; target rows fix the source as answer text+recent. Answer-text downweight scales source attention without adding it to image tokens; release-only removes the released mass from the source and renormalizes the row; task-only ranks image tokens by question--option relevance without the broad-attention penalty. Highlighted rows are the settings used by \method{}, and \(\Delta\) is the accuracy gain over CoT.}
\label{tab:module-ablation}
\end{table}

We then change one side at a time.
With \ve{} fixed as the target, answer text is stronger than recent-only, and answer text+recent gives the best result.
This matches the source design.
The answer text contains the base-selected option label and option string that the readout can reuse, while the recent prefix contains the reasoning that leads into the commitment.
With the source fixed as answer text+recent, all image patches and task-only visual relevance both improve over random patches, but neither reaches \ve{}.
Thus the target should not be all visual tokens, nor only patches close to the question-option context.
Removing patches that receive broad text attention gives a cleaner visual target.
The full setting reaches 53.8\% accuracy on matched MCQ rows.

\paragraph{Compute controls.}
We also compare methods at different inference budgets.
\method{} keeps the CoT trace fixed and uses its extra passes for evidence selection and final-answer readout.
The controls in Table~\ref{tab:cost-matched-controls} spend inference budget or intervention capacity on visual contrast, self-consistency, or generation-time visual steering.

\begin{table}[!t]
\centering
\small
\setlength{\tabcolsep}{2.0pt}
\begin{tabular}{lccc}
\toprule
Method & Compute & Acc. & \(\Delta\) \\
\midrule
\rowcolor{black!4}
CoT & \(1.0\times\) & 37.0 & -- \\
VCD & \(\sim 1.2\times\) & 40.5 & +3.5 \\
SC (\(k=2\)) & \(\sim 2.0\times\) & 40.0 & +3.0 \\
SC (\(k=4\)) & \(\sim 4.0\times\) & 40.7 & +3.7 \\
VISTA steering & \(\sim 1.0\times\) & 41.0 & +4.0 \\
\rowcolor{tverrow}
\textbf{\textcolor{tverpurple}{\method{}}} & \textcolor{tverpurple}{\(\bm{\sim 1.3\times}\)} & \textbf{\textcolor{tverpurple}{41.7}} & \textbf{\textcolor{tverpurple}{+4.7}} \\
\bottomrule
\end{tabular}
\caption{Compute controls on TreeBench. Cost is relative to one CoT and counts extra scoring, attention collection, or decoding; \(\Delta\) is over CoT.}
\label{tab:cost-matched-controls}
\end{table}

\method{} gives +4.7 points at about \(1.3\times\) compute, above the VCD (+3.5), self-consistency, and VISTA steering controls.
This indicates that the reallocation step matters, rather than the extra inference budget alone.

\begin{figure*}[!t]
\centering
\includegraphics[width=0.96\textwidth]{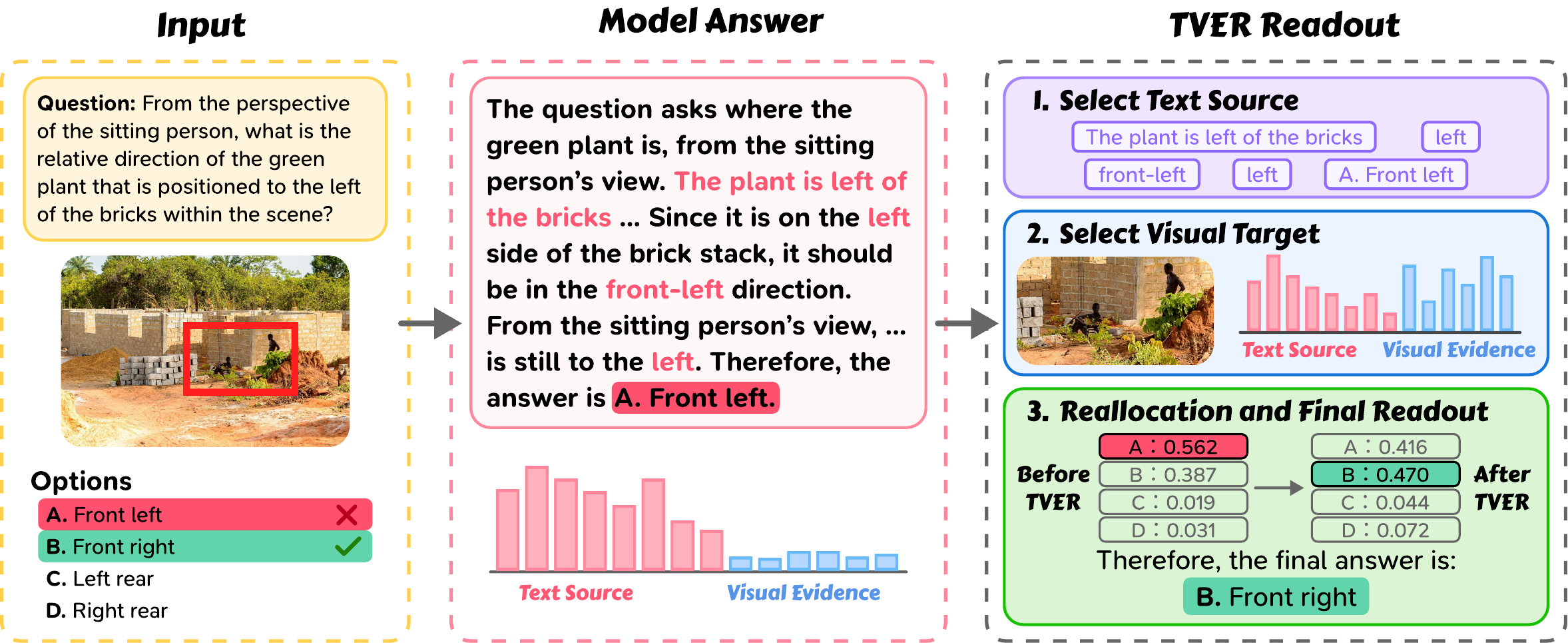}
\caption{Qualitative example on a TreeBench perspective question. The panels show the input, the original CoT, the selected text-side source and visual target, and the final score shift from A to B after \method{}.}
\label{fig:treebench-case}
\end{figure*}

\paragraph{Setting robustness.}
\label{sec:tver-sensitivity}
We next check whether the fixed setting is stable.
In Table~\ref{tab:tver-sensitivity}, \ve{} gives the best target set with 53.8\% accuracy.
Using 10\% or 50\% image tokens also helps, but \ve{} gives the largest gain.
The layer ablations follow the diagnostic in Figure~\ref{fig:diagnostic-trace}: middle-to-deep averaging is stronger than all-layer or single-window selector variants, and the same window is also best for intervention.
The transfer rows show a similar pattern.
A moderate transfer gives the best result, while moving too much mass weakens the gain.
We therefore keep \ve{}, middle-to-deep layers, \(\beta=0.6\), and \(\tau_{\max}=0.20\) for all main experiments.
Appendix~\ref{sec:backbone-sensitivity} further checks that this \(\beta\) setting is stable across all six backbones on TreeBench.

\begin{table}[!t]
\centering
\small
\setlength{\tabcolsep}{2.0pt}
\begin{tabular}{@{}llrr@{}}
\toprule
Factor & Setting & Acc. & \(\Delta\) \\
\midrule
\rowcolor{black!4}
Base & CoT & 50.6 & -- \\
\midrule
Target & VE-10\% & 53.1 & +2.5 \\
\rowcolor{tverrow}
Target & \textbf{\textcolor{tverpurple}{\ve{}}} & \textbf{\textcolor{tverpurple}{53.8}} & \textbf{\textcolor{tverpurple}{+3.2}} \\
Target & VE-50\% & 53.4 & +2.8 \\
\midrule
Selector & all-layer avg. & 53.4 & +2.8 \\
Selector & middle avg. & 53.5 & +2.9 \\
Selector & deep avg. & 53.2 & +2.6 \\
\rowcolor{tverrow}
Selector & \textbf{\textcolor{tverpurple}{middle-to-deep avg.}} & \textbf{\textcolor{tverpurple}{53.8}} & \textbf{\textcolor{tverpurple}{+3.2}} \\
\midrule
Window & shallow & 51.7 & +1.1 \\
Window & middle & 53.3 & +2.7 \\
\rowcolor{tverrow}
Window & \textbf{\textcolor{tverpurple}{middle-to-deep}} & \textbf{\textcolor{tverpurple}{53.8}} & \textbf{\textcolor{tverpurple}{+3.2}} \\
Window & deep & 53.1 & +2.5 \\
\midrule
Transfer & \(\beta=0.4,\tau_{\max}=0.20\) & 53.3 & +2.7 \\
\rowcolor{tverrow}
Transfer & \textcolor{tverpurple}{\(\bm{\beta=0.6,\tau_{\max}=0.20}\)} & \textbf{\textcolor{tverpurple}{53.8}} & \textbf{\textcolor{tverpurple}{+3.2}} \\
Transfer & \(\beta=0.8,\tau_{\max}=0.35\) & 53.6 & +3.0 \\
Transfer & \(\beta=1.0,\tau_{\max}=0.50\) & 53.0 & +2.4 \\
\bottomrule
\end{tabular}
\caption{Setting robustness on Qwen2.5-VL-7B tuning rows. Highlighted rows are final settings; \(\Delta\) is over CoT.}
\label{tab:tver-sensitivity}
\end{table}

\subsection{Qualitative Analysis}

Figure~\ref{fig:treebench-case} shows a perspective question.
The base CoT reads the scene but keeps using the scene-level relation that the plant is left of the bricks.
The question asks from the sitting person's view, so this repeated ``left'' cue pulls the final line toward the front-left option, even though the needed viewpoint evidence remains in the image.

\method{} acts at this readout step.
It uses the base-selected option words and their recent mentions as the text-side source, selects patches covering the plant, bricks, and sitting person, and moves final-answer attention toward this viewpoint evidence.
The CoT is not rewritten; only the answer line changes from the text-favored option to the image-supported option.

\FloatBarrier

\section{Related Work}

\begingroup
\setlength{\parskip}{0pt}
\noindent\textbf{Information-flow analysis.}
Visual-grounding studies have documented hallucination in image captioning, LVLM answers, and diagnostic benchmarks \citep{rohrbach2018objecthallucination,li2023pope,guan2024hallusionbench}.
These findings raise a more mechanistic question: how is visual evidence represented and routed through the model?
Recent work approaches this question through attention-gradient relevance, middle-layer signals, saliency changes, grounding heads, and information flow \citep{chefer2021generic,jiang2025attentionlens,zhang2026saliency,kang2025localizationheads,zhao2025infoflow}.
Our analysis builds on this line of work to examine visual evidence under a fixed CoT.
\par\vspace{0.5\baselineskip}
\noindent\textbf{Reasoning and decoding interventions.}
Existing methods intervene at several stages of multimodal reasoning.
Training and post-training methods improve trace generation through grounded CoT supervision or reinforcement-style optimization \citep{man2025argus,zhang2025improvevlmcot,liu2025visualrft,ni2025pointrft,huang2026visionr1,xie2026vigal}.
At inference time, some methods resample or search reasoning paths \citep{wang2023selfconsistency,wu2024vstar}.
Others steer generation using visual contrast, image attention, hidden states, or attention reallocation \citep{leng2024vcd,chen2024halc,liu2024pai,kang2025var,li2025vista,tu2026attnreal}.
Post-decoding methods instead revise the response after it has been generated \citep{zhou2024lure,yin2024woodpecker}.
\method{} keeps the CoT fixed and changes only the final-answer readout, making it complementary to methods that modify trace generation.
\par
\endgroup

\section{Conclusion}

In this paper, we propose Text-to-Visual Evidence Reallocation (\method{}) for final-answer repair in multimodal CoT under a fixed reasoning trace.
We find that a fixed trace can favor a wrong answer while useful visual evidence remains available at the final-answer readout.
Our method achieves consistent improvements across six open-source LVLMs and five benchmarks, covering both multiple-choice and open-answer evaluation.
When combined with generation-time steering, \method{} yields further gains, showing that final-answer repair can work alongside improvements to trace generation.

\section*{Limitations}

\noindent\textbf{1.} \method{} needs access to attention tensors during inference.
This is available for the open-source LVLMs used in our experiments, but API-only systems usually hide these states and may require a different implementation.

\noindent\textbf{2.} Our mechanism analysis also uses counterfactual visual lift and attention-gradient flow as diagnostic signals.
These probes help locate where visual evidence is available, but they should be read as correlational diagnostics rather than a full causal decomposition of the model's decision.

\noindent\textbf{3.} \method{} is designed for cases where useful visual evidence remains available at the final-answer readout.
It is not expected to repair traces whose needed evidence has already been lost, whose image evidence is absent or ambiguous, or whose error comes from non-visual reasoning outside the selected target.

\begingroup
\emergencystretch=2em
\bibliography{references}
\endgroup

\appendix

\section{Mechanism Diagnostics}
\label{sec:additional-evidence}

In this section, we further analyze how \method{} changes the fixed-trace readout and whether visual support for the correct answer remains in CoT error cases.
Note that all diagnostics are computed offline after evaluation, and \method{} does not use ground-truth labels at inference time.

\subsection{Repair and Damage}

\begin{center}
\footnotesize
\setlength{\tabcolsep}{2.2pt}
\begin{tabular}{@{}lccrr@{}}
\toprule
Bench. & $\Delta$ & 95\% CI & Repair & Damage \\
\midrule
Tree & +4.69 & [+2.22, +7.16] & 9.0\% & 2.7\% \\
V* & +3.14 & [+0.00, +6.28] & 17.8\% & 1.4\% \\
Math & +1.80 & [+0.60, +3.10] & 9.1\% & 1.6\% \\
MMStar & +2.60 & [+1.60, +3.67] & 9.6\% & 1.4\% \\
RH-Reason & +3.00 & [+0.33, +5.67] & 7.2\% & 3.4\% \\
\bottomrule
\end{tabular}
\apptablecaption{tab:repair-damage}{Paired accuracy gains and answer transitions for Qwen2.5-VL-7B. $\Delta$ and 95\% CIs use pre-rounding counts and are reported in percentage points. Repair is wrong-to-correct among CoT-wrong examples; damage is correct-to-wrong among CoT-correct examples.}
\end{center}

\subsection{Closed-Set Rescoring}
\label{sec:diagnostic-prevalence}

We further evaluate \method{} by ranking all candidate options under the same fixed CoT, rather than decoding the final answer as in the main evaluation.
In Table~\ref{tab:closedset-appendix}, \method{} improves top-1 option accuracy on all five pools (54.8 to 58.5 on average) and increases the ground-truth margin over the strongest wrong option by .066, indicating stronger relative support for the correct option at the final readout.

\begin{center}
\footnotesize
\setlength{\tabcolsep}{2.4pt}
\begin{tabular}{@{}lrrrr@{}}
\toprule
Bench. & CoT & \method{} & \(\Delta\) & Margin \\
\midrule
Tree & 42.0 & 46.7 & +4.7 & +.083 \\
V* & 54.5 & 58.1 & +3.7 & +.074 \\
Math & 65.0 & 67.8 & +2.8 & +.052 \\
MMStar & 60.7 & 64.0 & +3.3 & +.061 \\
RH-MCQ & 52.0 & 55.7 & +3.7 & +.062 \\
\midrule
Average & 54.8 & 58.5 & +3.6 & +.066 \\
\bottomrule
\end{tabular}
\apptablecaption{tab:closedset-appendix}{Closed-set rescoring on Qwen2.5-VL-7B using all multiple-choice items from the five benchmark releases. Margin is the average increase in the score gap between the ground-truth option and the strongest wrong option. The \(\Delta\) values are computed from unrounded scores.}
\end{center}

\subsection{Hidden-State Visual Lift}

This diagnostic asks whether CoT error rows still contain image-conditioned support for the ground-truth option.

Let \(s^\ell(o\mid Q,I',c_{\leq t})\) be the option score obtained by applying the LM readout head to the layer-\(\ell\) hidden state at the answer-readout position, using the same fixed prefix and option-scoring template as \(s(o\mid Q,I',c_{\leq t})\).
We define the hidden-state visual lift as
\begin{equation}
\begin{aligned}
H_t^\ell(o)=
&s^\ell(o\mid Q,I,c_{\leq t})\\
&-\frac{1}{|\mathcal{C}|}\sum_{\tilde I\in\mathcal{C}}
s^\ell(o\mid Q,\tilde I,c_{\leq t}).
\end{aligned}
\end{equation}
We average \(H_t^\ell(o)\) over the same middle-to-deep readout window used by \method{} and compute the correct-minus-selected lift gap on base-CoT error rows:
\begin{equation}
\Delta^{\mathrm{lift}} =
\mathbb{E}_{\ell\in\mathcal{L}}\left[
H_t^\ell(o_{\mathrm{gt}})-H_t^\ell(o_{\mathrm{sel}})
\right].
\end{equation}
Positive values mean that, after visual counterfactual subtraction, the hidden state assigns stronger image-conditioned support to the correct option than to the selected wrong option.
Table~\ref{tab:hidden-lift-appendix} reports the diagnostic result.
The lift gap is positive on every closed-set pool.
The pattern fits the main account: \method{} helps when usable correct-answer visual evidence is already present in the readout states but is underweighted by the final answer posterior.

\begin{center}
\footnotesize
\setlength{\tabcolsep}{2.1pt}
\begin{tabular}{@{}lrr@{}}
\toprule
Bench. & \(\Delta^{\mathrm{lift}}\) & Correct+ \\
\midrule
Tree & +.071 & 64.3\% \\
V* & +.064 & 63.2\% \\
Math & +.041 & 57.7\% \\
MMStar & +.053 & 60.1\% \\
RH-MCQ & +.056 & 61.1\% \\
\midrule
Average & +.057 & 61.3\% \\
\bottomrule
\end{tabular}
\apptablecaption{tab:hidden-lift-appendix}{Hidden-state visual lift diagnostics on CoT error rows. \(\Delta^{\mathrm{lift}}\) is the correct-minus-selected lift gap averaged over the middle-to-deep readout window. Correct+ is the share of rows with a positive gap.}
\end{center}

\section{Experimental Details}

\subsection{Backbone Sensitivity}
\label{sec:backbone-sensitivity}

The main text selects \(\beta=0.6\) on Qwen2.5-VL-7B tuning rows.
We further test this setting across all six backbones on TreeBench, while keeping the visual target, layer window, and cap fixed.

\begin{center}
\footnotesize
\setlength{\tabcolsep}{2.0pt}
\begin{tabular}{@{}lcccc@{}}
\toprule
Model & \(\beta_{\mathrm{low}}\) & \(\beta^\star\) & \(\beta_{\mathrm{high}}\) & Range \\
\midrule
Qwen2.5-VL-7B & 0.4 & 0.6 & 0.8 & 40.0--41.7 \\
Qwen2.5-VL-32B & 0.4 & 0.6 & 0.8 & 43.5--44.9 \\
InternVL3-8B & 0.4 & 0.6 & 0.8 & 41.7--42.7 \\
InternVL3-38B & 0.4 & 0.6 & 0.8 & 43.7--44.9 \\
LLaVA-OV-7B & 0.4 & 0.6 & 0.8 & 40.7--41.7 \\
LLaVA-OV-72B & 0.4 & 0.6 & 0.8 & 42.7--44.0 \\
\bottomrule
\end{tabular}
\apptablecaption{tab:backbone-sensitivity}{Backbone sensitivity around the fixed \(\beta=0.6\) setting on TreeBench. The visual target, layer window, and cap are kept at the main \method{} setting.}
\end{center}

As shown in Table~\ref{tab:backbone-sensitivity}, \(\beta=0.6\) gives the best result for every backbone, while the adjacent settings remain within 1.7 points.

\end{document}